\documentclass[runningheads]{llncs}

\usepackage{amsmath,amsthm}

\usepackage[table]{xcolor}
\definecolor{lg}{gray}{0.9}
\usepackage{multirow}
\usepackage[T1]{fontenc}
\usepackage{graphicx}
\usepackage{makecell}
\usepackage{xcolor}

\usepackage[ruled,vlined]{algorithm2e}
\usepackage{tikz}
\usepackage{subcaption}
\usepackage{appendix}
\usepackage{lscape}
\usepackage[figuresleft]{rotating}
\newcommand*{\doi}[1]{\href{http://dx.doi.org/#1}{doi: #1}}

\usepackage{epstopdf}
\begin{document}
\title{Explainability of Predictive Process Monitoring Results: Can You See My Data Issues?}
\titlerunning{Explainability of Predictive Process Monitoring results}
%

\author{Ghada Elkhawaga\inst{1,2}\orcidID{0000-0001-7801-7310} \and
Mervat Abuelkheir\inst{3}\orcidID{0000-0003-0958-7322} \and
Manfred Reichert \inst{1}\orcidID{0000-0003-2536-4153}}
\authorrunning{G. Elkhawaga et al.}
%
\institute{Institute for Databases and Information Systems, Ulm University, Ulm, Germany \\ \email{\{ghada.el-khawaga,manfred.reichert\}@uni-ulm.de}\\
\and
 Faculty of Computers and Information, Mansoura University, Dakahlia, Egypt
\and
Faculty of Media Engineering and Technology, German University in Cairo, New~Cairo, Egypt\\
\email{ mervat.abuelkheir@guc.edu.eg}}
\maketitle              
\begin{abstract}
Predictive business process monitoring (PPM) has been around for several years as a use case of process mining. PPM enables foreseeing the future of a business process through predicting relevant information about how a running process instance might end, related performance indicators, and other predictable aspects. A big share of PPM approaches adopts a Machine Learning (ML) technique to address a prediction task, especially non-process-aware PPM approaches. Consequently, PPM inherits the challenges faced by ML approaches. One of these challenges concerns the need to gain user trust in the predictions generated. The field of explainable artificial intelligence (XAI) addresses this issue. However, the choices made, and the techniques employed in a PPM task, in addition to ML model characteristics, influence resulting explanations. A comparison of the influence of different settings on the generated explanations is missing. To address this gap, we investigate the effect of different PPM settings on resulting data fed into a ML model and consequently to a XAI method. We study how differences in resulting explanations may indicate several issues in underlying data. We construct a framework for our experiments including different settings at each stage of PPM with XAI integrated as a fundamental part. Our experiments reveal several inconsistencies, as well as agreements, between data characteristics (and hence expectations about these data), important data used by the ML model as a result of querying it, and explanations of predictions of the investigated ML model. 

\end{abstract}

\section{Introduction} \label{sec1}
\subsection{Problem Statement}
Process mining is a scientific field at the intersection between Business Process Management (BPM) and data science. Process mining enables contrasting the ideal view of how process activities shall be carried out, i.e., the normative model of a business process, with how process activities are actually carried out, i.e., the descriptive model of a business process \cite{PMBook}.
Predictive process monitoring (PPM) \cite{b2}, as a practical use case of process mining, supports stakeholders with predictions about the future of a running business process instance. A business process \cite{fundametalBPM} represents a sequence of executed events affected by decision points and involving a number of actors to achieve one or more outcomes representing goals of the business process. Business process-related predictions may enable stakeholders to take preventive decisions in case of undesired outcomes or expected delays or resource congestion.

As stakeholders engagement is at the center of process mining tasks, performance and accuracy are not the only aspects which matter while carrying on a PPM prediction task. While depending on ML models in predicting the future of running business process instances, it becomes necessary to persuade business stakeholders of the validity of reasoning mechanisms followed by a predictive model. Justifying predictions to their recipients enable gaining users' trust, engagement and advocacy of PPM employed mechanisms.

To this end, eXplainable Artificial Intelligence (XAI) \cite{imlbook} methods and mechanisms \cite{shapL,PDP,DeepLIFT,LIMEPap,ALE,LRP,Modelagnostic} are put in place to provide explanations of predictions generated by a ML model. These explanations are obtained either in parallel to the prediction process or afterwards. However, the aforementioned explanations are expected to reflect how a predictive model is influenced by different choices made through the ML pipeline. Such pipeline starts with data analysis and cleaning, data preprocessing, model choosing, parameters selection and model predictions evaluation. Moreover, PPM tasks employ specific mechanisms to ensure aligning process mining artefacts to ML models requirements. Therefore, while having different XAI methods addressing explainability needs in the context of PPM, studying the effects of different PPM workflow settings should be considered as an enabling step to evaluate explainability outcomes.

\subsection{Contributions}
With an attempt to address the need to understand and gain insights into the application of XAI methods into PPM, this paper presents: 
\begin{itemize}
   \item {A study of the effect of underlying choices made in PPM task and how this effect can revealed through explanations generated by different XAI methods.}
    \item {A study of explanations generated by two different global XAI methods, and two model-specific methods, for predictions of two predictive models over process instances from 29 event logs preprocessed with two different preprocessing combinations,}
    \item {As a major contribution, this work provides an open access framework of various XAI methods built upon different PPM workflow settings.}
\end{itemize}

Section \ref{Section 2} provides background information on basic topics needed for understanding this work. In Section \ref{Section 3}, we highlight the basic research questions investigated in this paper. In Sections \ref{5} and \ref{6}, we discuss the settings of the conducted experiments, experimental results and observations. Section \ref{Section 7} highlights lessons learned and conclusions answering the basic research questions. Related work is illustrated in Section \ref{Section 8}. Finally, we conclude the paper in Section \ref{Section 10}.

\section{Backgrounds} \label{Section 2}
This section introduces basic concepts and background knowledge necessary to understand our work. Section \ref{PPMsec} introduces predictive monitoring with a focus on outcome-oriented predictions. Then, we discuss available explainability methods with an in-depth look into those addressing tabular data as these data type is the focus of this work.

\subsection{Predictive Process Monitoring} \label{PPMsec}
 Predictive Process Monitoring (PPM) is a subfield of BPM and a process mining use case that provides decision makers with predictions of the future of a running process instance. This goal can be realised by building models to generate predictions about running process-related information, e.g. the next activity to be carried out, time-related information (e.g. elapsed time, remaining time till the end), outcome of the process instance, execution cost, or executing resource \cite{PMBook}. Generating such predictions is what we denote as PPM tasks. A PPM task takes an event log as one of its inputs. An event log documents the execution history of a process terms of traces, each of them representing execution data belonging to a single process instance. 
\theoremstyle{definition}
\begin{definition}[Trace, event log]
Let $\varepsilon$  be the set of all possible events taking place in a business process, i.e., the event universe. A trace $\sigma$ is a sequence events $\sigma = <e_{1},e_{2},e_{3},...,e_{n}>$ $\in$ $\varepsilon$ $\forall$ n $\in$ N being the total number of events in the trace. A trace contains at least one event. Let D be the universe of data attributes associated with each event. $d_{ij}$ $\in$ D, where i is the event number, and j $\in$ M being the number of data attributes associated with the event.\\ A trace $\sigma$ = $< (e_{1},(d_{11},d_{12},...,d_{1m})),(e_{2},(d_{21},d_{22},...,d_{2m})),(e_{3},(d_{31},d_{32},...,d_{3m}))\\,...,(e_{n},(d_{n1},d_{n2},...,d_{nm}))>$. An event with its associated attributes can not take place more than once in a trace, i.e., $e_{x},(d_{x1},d_{X2},...,d_{Xm}) \neq e_{y},(d_{y1},d_{y2},...,d_{ym})$.
An event log (L) corresponds to a set of traces, where each event appears at most once in the entire event log \cite{PMBook}. 
\end{definition}
As mandatory attributes, a trace should contain \emph{case ID} as a unique identifier of the trace, \emph{event class} which represents a step carried out in fulfillign th process instance, and the \emph{timestamp} denoting the time the event took place \cite{PMBook}.

A trace may contain optional data items representing information about a single event. These items are called data attributes or data payload. Sometimes they are also denoted as \emph{dynamic attributes} since they have different values for each event in a trace. A single case may be represented by several records, each represents a single event along with its data payload. Besides dynamic attributes there are static ones associated with each trace. \emph{Static attributes} have constant values for all events of a given trace. Note that we use the terms case, trace, and process instance interchangeably to refer to the same concept. However,we tend to place the term \emph{case} where the data view of a business process is our concern. We use the term \emph{trace} in the context concerned with the logical view of a business process. Finally, we use the term \emph{process instance} when we are concerned with the conceptual view of a business process. 

\subsubsection{PPM Workflow}
\begin{sidewaysfigure}
\centering
\includegraphics[width=1\textwidth, height=0.6\textheight]{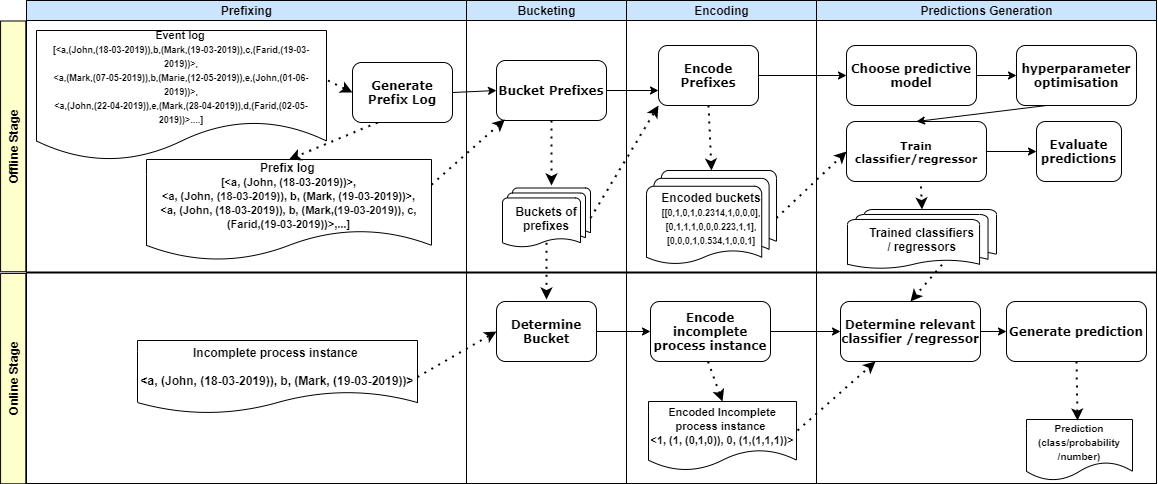}
\caption{PPM Workflow} \label{PPMWF}
\end{sidewaysfigure}

According to the survey results reported in \cite{b2}, a PPM task follows a workflow of four steps, which are organised in two stages, an offline and another is online. Each PPM task starts with an event log being preprocessed and processed offline. With offline preprocessing and processing, we mean transforming the historical traces of a business process stored in the form of an event log and using them to train a predictive model. The outcome of this preprocessing is passed to another stage, where an incomplete process instance is processed online, i.e., at runtime. The complete workflow is shown in Figure \ref{PPMWF} and discussed in the following subsection.

\begin{enumerate}
    \item {\textbf{PPM offline Stage.}} An event log is input to this stage. Usually, this event log satisfies process mining requirements, meaning that it is organised in the form of traces. Each trace has certain mandatory and optional attributes. Preprocessing steps are then carried out to transform traces in a form compatible with constraints and requirements of the prediction generation process.

\begin{enumerate}
    \item {\textbf{Prefix log construction.}} As stated in \cite{b2}, most contemporary PPM approaches that use ML, take as input a prefix log constructed from the input event log. Note that a predictive model is expected to predict information related to an incomplete process instance (i.e., a partial trace), so it has to be trained on incomplete versions of historical process instances kept in the event log. The tendency of a predictive model to be biased towards patterns that were learned from longer process instances is another reason that necessitates the use of prefixes instead of complete process instances. According to \cite{b2}, prefixes generated from an event log can increase to the extent of slowing down the overall prediction process. Therefore, prefixes can be generated by truncating a process instance up to a predefined number of events. As proposed in \cite{clus}, truncating a process instance can be done up to the first k events of its trace, or up to k events with a gap (g) step separating each two events, where k and g are user-defined. The latter prefixing approach is called \emph{gap-based prefixing}.

\begin{definition} [Prefix trace, prefix log] \\
 Let $\sigma$ = $< e_{1},(d_{11},...,d_{1m}),e_{2},(d_{21},...,e_{n},(d_{n1},...,d_{nm})>$ be a trace with n $\in$ N being the number of events and m $\in$ M being the number of event attributes. A prefix $P_{i}$ generated from $\sigma$ = $P_{i-1} \cup <e_{1+(i-1)*g}>$, where $P_{1}=<e_{1}>$, g is the gap,  $1 < i \leq N$  and $1 \leq g \leq N-1$. A prefix log is the set of all prefixes generated from each trace in the event log under preprocessing.   
\end{definition}

After obtaining a prefix log from the traces in a given event log, the prefixes need to be further preprocessed in order to serve as appropriate input for a predictive model. Prefix preprocessing steps include bucketing and encoding.

\item{\textbf{Bucketing.}}
Prefix bucketing means grouping the prefixes according to certain criteria (e.g., number of activities or reaching a certain state in the process execution). Prefixes in the same bucket are treated as a unit during the encoding and training steps in the offline stage. They also represent a unit during the online stage while comparing an incomplete process instance to find its relevant group of instances, i.e., bucket. Bucketing techniques can be state-based, clustering, single, domain knowledge-based, or prefix-length-based. For example, in \emph{single} bucketing, all prefixes generated from the traces of an event log are treated as a single bucket. Meanwhile in \emph{prefix-length-based} bucketing, as the name indicates, prefixes of the same length are bucketed together. Afterwards, a separate predictive model is created for each bucket of prefixes.

\item{\textbf{Encoding.}}
Prefix encoding means transforming a prefix to a feature vector that serves as input to the predictive model, either for training or for making predictions. A predictive model can only handle numerical values. Therefore, all categorical attributes of a training dataset or a training sample need to be first transformed into a numerical form. The same applies to process instances whose prefixes serve as input for a predictive model. In the context of PPM encoding, a slight change of the regular ML encoding step is involved. As reviewed in \cite{b2}, \emph{Static encoding} technique is used to encode static attributes of a prefix event log. Meanwhile, aggregation, index-based and last state are all examples of encoding techniques used to encode dynamic attributes of prefix event logs. Different encoding techniques yield different sizes of encoded prefixes with different types of included information. Such diversity is proven to affect both the accuracy of the predictions and the efficiency of the prediction process in terms of execution times and needed resources \cite{b2}.

\item{\textbf{Predictive model construction and operation.}} Depending on the PPM task, the respective predictive model is chosen. The prediction task type is not the only factor guiding a predictive model selection process. Other factors include scalability of the model when facing larger amounts of data, simplicity and interpretability of results. \cite{b2} claimed that Decision Trees (DT)s are dominating as a choice in most current research on PPM, due to their simplicity. XGBoost represent another type of outperforming predictive models used in PPM tasks \cite{b2}.

Parameters values assignment follows the model selection step. The values of model parameters are learned by the model during the training phase, whereas the values of hyperparameters are set before training the model. Next step is to train the predictive model on encoded prefixes representing completed process instances. Note that for each bucket a dedicated predictive model needs to be trained, i.e., the number of predictive models depends on the bucketing technique chosen. After generating predictions for the training dataset, the performance of a predictive model needs to be evaluated. The choice of respective evaluation technique depends on the prediction task, i.e., classification tasks have different evaluation metrics than regression tasks.\\
\end{enumerate}

\item{\textbf{PPM online Stage.}} This stage starts with an incomplete process instance, i.e., a running process instance. Buckets formed in the offline stage are recalled to determine the suitable one for the running process instance. This is accomplished based on the similarity between the running process instance and prefixes in a bucket according to the criteria defined by the bucketing method. Afterwards, the running process instance is encoded according to the encoding method chosen for the PPM task. The encoded form of the running process instance becomes qualified as an input to the prediction method after determining the relevant predictive model from the models created in the offline stage. Finally, the online stage is concluded by the predictive model generating a prediction for the running process instance according to the predefined goal of the PPM task.
\end{enumerate}
\subsection{eXplainable Artificial Intelligence} \label{XAI}

Explainability, interpretability, and transparency are common terms with more or less the same meaning, referring to the problem of understanding and trusting the underlying mechanisms as well as the predictions of a ML model \cite{EvalQual}. According to \cite{accountAI}, an explanation is \emph{"a human-interpretable description of the process by which a decision maker took a particular set of inputs and reached a particular conclusion"}. In the context of our research, the decision maker is a ML-based predictive model. Moreover, an explanation has a broader meaning including not only the reasoning process behind the generation of predictions, but also the factors contributing to reach a prediction. For example, it is crucial to know which data characteristics are considered deterministic of a prediction, and which features are considered important to the predictive model. 

Interpretability is \emph{the degree to which a human can consistently predict the model's result} \cite{imlbook}. This human prediction and the mimicking of the model's reasoning is based on the mental model a human forms with respect to how the model reached its decisions. In \cite{imlbook,XAIResp}, interpretability is considered being equal to transparency. Transparency can be regarded through three levels. The first is \emph{simulatability} which refers to the human's capability to simulate how a model reaches a prediction. Another level is \emph{decomposability}, meaning understanding a predictive model in terms of its elementary components, e.g., its inputs and hyperparameters. The last aspect is \emph{algorithmic transparency}, denoting the understandability of a prediction algorithm inner working \cite{imlbook}.

Model transparency can be an inherent characteristic of itself or be achieved through an explanation of a model. Transparent models are understandable on their own and satisfy one or all model transparency levels \cite{XAIResp}. Linear models, decision trees, Bayesian models, rule-based learning, and General Additive Models (GAM) are all considered being transparent/interpretable models varying in degree of transparency achievement.  

However, the mental model of humans and their ability to mimic the reasoning process of a predictive model and, hence, to trust and accept its predictions depend on certain qualities of a good explanation. The composition, content and quality of an explanation can be influenced, characterised and then evaluated according to its constructing approach. XAI methods can be differentiated according to several factors. These factors include how an explanation is being generated, the granularity and the scope to which an explanation can be generalised, when to apply a XAI method, how an explanation is presented, and the user group addressed by an explanation. 

Generating explanations can be done using several ways including, explanations by providing the user with \emph{examples of process instances} which are similar in their attributes while being different in the prediction. Another way is to \emph{visualise} intermediate representations and layers of a predictive model with the aim to qualitatively determine what a model has learned \cite{EvalQual}.  \emph{Explanation by feature relevance} includes assigning importance scores to features contributing to a ML model prediction represent one type of XAI methods. The latter include SHAP \cite{shapL} and partial dependence plot (PDP) \cite{PDP} as examples. An explanation can be generated \emph{locally} for one process instance or a sample, or may be applied \emph{globally} to the reasoning process of a predictive model over the complete event log or dataset. Accumulated Local Effects (ALE) \cite{ALE} is an example of global XAI methods, while LIME \cite{LIMEPap} is considered a local XAI method.

Choosing the form an explanation is presented is determined by the way the explanation is generated, the characteristics of the end user (e.g., level of expertise), and the scope of the explanation, the purpose of generating an explanation, (e.g., to visualise effects of feature interactions on decisions of the respective predictive model). \cite{multidiscip} introduces three categories of presentation forms including \emph{visual}, \emph{verbal}, and \emph{analytic} explanations.

The point of time to generate an explanation is another determining factor of a XAI method. The depth at which an explanation is able to reflect a predictive model's reasoning process internally or to find a mapping between the model's inputs and its outputs determines whether to rely on \emph{intrinsic explanations} or \emph{post-hoc} ones. When explainability is imposed through mechanisms put in place while constructing the model to obtain a white-box predictive model, it is called intrinsic explanations. Meanwhile, using an explanation method to understand the reasoning process of a  model in terms of its outcomes is called \emph{post-hoc explanation} of a predictive model.

XAI methods are further categorised as being either \emph{model-agnostic} or \emph{model-specific}. Model-agnostic approaches are able to explain any type of ML predictive model, whereas model-specific approaches can only be used on top of specific models. For example, DeepLift \cite{DeepLIFT} and LRP \cite{LRP} provide explanations for neural network-based predictive models. \cite{Modelagnostic} discusses advantages of model-agnostic approaches focusing on the flexibility of model choice, explanation form and representation. In turn, \cite{PrincPract} argues against the flexibility of model-agnostic approaches highlighting that these approaches make assumptions about explained predictive models to maintain the expected flexibility. 

\section{Research Questions}\label{Section 3} 
The goal of this research is to study how data preprocessing choices made in the PPM workflow can be reflected in retrieved explanations of predictions. It is also desired to study how the former choices affect the ability of XAI methods to result in meaningful explanations. By meaningful explanations we mean those that enable identifying useful patterns to the predictive model. As illustrated in Section \ref{XAI}, decomposability constitutes an essential component of achieving transparency. Therefore, it is crucial to ensure that an explanation allows tracing back a prediction to inputs, their characteristics, phenomena related to these characteristics, and the choices made through out the process of prediction generation, e.g., encoding and bucketing techniques-related choices. We need to study how explanations of different XAI methods be affected by characteristics and sensitivities of the explained predictive model. Overall, this leads to the following research questions (RQ)s:

\textbf{RQ1: To what extent can an explanation reflect the effects of dataset characteristics on predictive model reasoning process?} We want to study to what extent horizontal and vertical characteristics of a dataset are reflected by explanations on PPM results. With \emph{horizontal characteristics}, we mean dataset size and imbalanced class ratios. Whereas, \emph{vertical characteristics} refers to correlations and dependencies between features that may affect conclusions made by explainability techniques. A predictive model learns patterns from data and these patterns are believed to inherit biases that might be present in the data, resulting in a biased model. To understand to what extent an explanation is sensitive to data biases and imbalances, is the ultimate goal of RQ1.

\textbf{RQ2: What is the implication of a certain preprocessing combination on a XAI method?} As illustrated in Section \ref{PPMsec}, different preprocessing approaches may be combined to transform process instances into form suited as input for a predictive model. Different combinations of bucketing and encoding techniques result in different prefixes, which vary in size and information content. Prefixes are bucketed in different buckets which group them according to different criteria, and hence, these buckets differ in size and information content. Complex inputs sometimes manipulate given characteristics of certain predictive models to cope with emerging input complexity. For example, linear models are transparent by nature \cite{XAIResp}, i.e., they are simulateable, decomposable and algorithmically transparent. However, as confirmed by many studies \cite{mythos,XAIResp}, even transparent models (including linear ones) lose their simulatability and decomposability at the presence of high dimensional or heavily engineered features. Therefore, preprocessing approaches, which increase dimensionality of the feature set, may influence the explainability of certain predictive models.  

\textbf{RQ3: How can the self-transparency of a ML model affect a XAI generated explanation?} Predictive models also differ with respect to transparency and accuracy levels. According to \cite{XAIResp}, it is stated that a trade-off exists between model performance and transparency. Furthermore, transparent models that approximate a complex model performance in a certain vicinity can not be generalised \cite{PrincPract}. How explanations of predictions generated by models of different transparency and complexity levels may differ is one of the targets of this research. Moreover, self-explanatory mechanisms provided by the predictive model itself have to be differentiated in terms of consistency over several runs of the model. Consistency means giving the same feature set the same weights after querying the model for the important features several times. 
\section{Experiments} \label{5}
This section describes choices of the experiments we performed to answer research questions in Section \ref{Section 3}. For the basic infrastructure of a PPM outcome prediction task, we are inspired by the framework and findings demonstrated in \cite{b2} and available at \cite{b2git}. Note that we are not changing any of the steps carried out through the aforementioned framework. Settings preservation is needed to be able to observe the impact of the settings described in \cite{b2}, given the reported performance of studied predictive models and preprocessing techniques, from an explainability perspective.

\begin{figure}
\centering
\includegraphics[width=0.8\textwidth, height=6.5 cm]{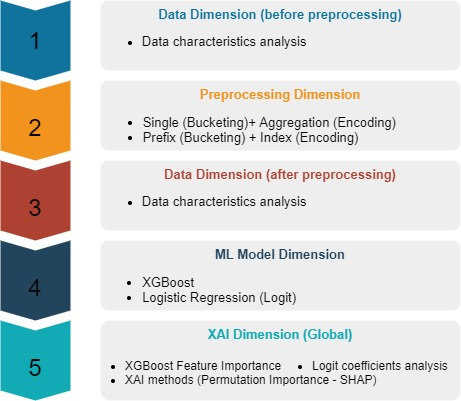}
\caption{Experiments Taxonomy} \label{ExpTax}
\end{figure}

Figure \ref{ExpTax} shows the taxonomy of the implemented experiments organized under dimensions resembling a ML model creation pipeline, being aligned with the PPM offline workflow, and incorporating an explainability-related dimension. These dimensions are a means to categorise the experiments with the aim of answering our research questions. These dimensions are further discussed through this section. All experiments were run using Python 3.6 and the scikit-learn library \cite{scikitlearn} on a 96 core of a Intel(R) Xeon(R) Platinum 8268 @2.90GHz with 768GB of RAM. Note that we apply all available combinations from each dimension, using the taxonomy defined in Figure \ref{ExpTax} in a dedicated experiment while fixing other options from other dimensions. The code of executed experiments is available through our Github repository\footnotemark[1] \footnotetext[1]{\url{https://github.com/GhadaElkhawaga/PPM\_XAI\_Comparison.git}}to enable open access for interested practitioners. 
  
\subsection{Experimental setup} \label{ExpSetup}
In this subsection, we describe the building blocks of our experiments, including data, chosen preprocessing techniques, selected predictive models, and employed XAI algorithms. Categorising our experiments, and the techniques we employed (as shown in Figure \ref{ExpTax}) is done with the aim of studying the impact of choices on resulting explanations. Observing the impact of modifying different parameters at each dimension is planed to the be the basis of this study.
\subsubsection{Data dimension} 
The experiments are carried on four real-life event logs which are publicly available from the 4TU Centre for Research Data \cite{4TU}. The chosen event logs vary in the considered domain (healthcare, government and banking), the number of traces (representing process instances), and the number of events in each trace. The chosen event logs further vary also in the number of static and dynamic attributes, the number of categorical attributes and, as a result, the number of categorical levels available through each categorical attribute. 

Choices in this dimension are mainly impacted by the need to study issues formulated in \textbf{RQ1}.The four basic event logs \cite{4TU} used are as follows:
\begin{itemize}
    \item {\textbf{Sepsis.}} This event log belongs to the healthcare domain and reports cases of Sepsis as a life threatening condition.
    \item {\textbf{Traffic fines.}} This event log is a governmental one extracted from an Italian information system for managing road traffic fines.
    \item {\textbf{BPIC2017.}} This event log documents load application process in a Dutch financial institution. 
    \item {\textbf{Hospital billing.}} This event log is extracted from an ERP system of a hospital. It documents the billing procedure for medical services.
\end{itemize}

\cite{b2} applies different labelling functions to classify each process instance into one of two classes, i.e., a binary classification task. These labelling functions result in three extracted logs from the \emph{Sepsis} event log, three logs extracted from \emph{BPIC2017} event log, two logs extracted from \emph{Hospital\_billing} log. These different labelling functions increased the number of used event logs from four to nine event logs.

We have chosen event logs with different positive class ratios, varying between <1\% and 47\% to study the effects of imbalanced event logs on explanations of different granularity. Table \ref{statistics} shows  basic statistics of the event logs used in our experiments. These event logs are cleaned and transformed applying the same rules suggested by the framework available in \cite{b2}. Moreover, the used labelling functions defining how to label a trace based on a certain condition are defined in \cite{b2}. 
\begin{table}
\caption{Event logs statistics.}\label{statistics}
\begin{tabular}{|l|l|l|l|l|l|l|l|l|l|l|l|l|l|l|l|}
\hline
 \tiny{\thead{Event\\log}}&\tiny{\thead{\#\\traces}}&\tiny{\thead{Short.\\trace\\len.}}&\tiny{\thead{Avg.\\trace\\len.}}&\tiny{\thead{Long.\\trace}}&\tiny{\thead{Max\\prfx\\len.}}&\tiny{\thead{\#\\trace\\variants}}&\tiny{\thead{\%pos\\class}}&\tiny{\thead{\#\\event\\class}}&\tiny{\thead{\#\\static\\col}}&\tiny{\thead{\#\\dynamic\\cols}}&\tiny{\thead{\#\\cat\\cols}}&\tiny{\thead{\#\\num\\cols}}&\tiny{\thead{\#\\cat\\levels\\static\\cols}}&\tiny{\thead{\#\\cat\\levels\\dynamic\\cols}}\\
\hline
\tiny Sepsis1&\scriptsize 776&\scriptsize 5&\scriptsize 14&\scriptsize 185 &\scriptsize 20&\scriptsize 703&\scriptsize 0.0026&\scriptsize 14&\scriptsize 24&\scriptsize 13&\scriptsize 28&\scriptsize 14&\scriptsize 76&\scriptsize 38\\ \hline
\tiny Sepsis2&\scriptsize 776&\scriptsize 4&\scriptsize 13&\scriptsize 60&\scriptsize 13&\scriptsize 650&\scriptsize 0.14&\scriptsize 14&\scriptsize 24&\scriptsize 13&\scriptsize 28&\scriptsize 14&\scriptsize 76&\scriptsize 39\\ \hline
\tiny Sepsis3&\scriptsize 776&\scriptsize 4&\scriptsize 13&\scriptsize 185&\scriptsize 31&\scriptsize 703&\scriptsize 0.14&\scriptsize 14&\scriptsize 24&\scriptsize 13&\scriptsize 28&\scriptsize 14&\scriptsize 76&\scriptsize 39\\ \hline
\tiny{\makecell{Traffic\\fines}}&\scriptsize 129615&\scriptsize 2&\scriptsize 4&\scriptsize 20&\scriptsize 10&\scriptsize 185&\scriptsize 0.455&\scriptsize 10&\scriptsize 4&\scriptsize 14&\scriptsize 13&\scriptsize 11&\scriptsize 54&\scriptsize 173\\ \hline
\tiny{\makecell{BPIC2017\_\\Accepted}}&\scriptsize 31413&\scriptsize 10&\scriptsize 35&\scriptsize 180&\scriptsize 20&\scriptsize 2087&\scriptsize 0.41&\scriptsize 26&\scriptsize 3&\scriptsize 20&\scriptsize 12&\scriptsize 13&\scriptsize 6&\scriptsize 682\\ \hline
\tiny{\makecell{BPIC2017\_\\Cancelled}}&\scriptsize 31413&\scriptsize 10&\scriptsize 35&\scriptsize 180&\scriptsize 20&\scriptsize 2087&\scriptsize 0.47&\scriptsize 26&\scriptsize 3&\scriptsize 20&\scriptsize 12&\scriptsize 13&\scriptsize 6&\scriptsize 682\\ \hline
\tiny{\makecell{BPIC2017\_\\Refused}}&\scriptsize 31413&\scriptsize 10&\scriptsize 35&\scriptsize 180&\scriptsize 20&\scriptsize 2087&\scriptsize 0.12&\scriptsize 26&\scriptsize 3&\scriptsize 20&\scriptsize 12&\scriptsize 13&\scriptsize 6&\scriptsize 682\\ \hline
\tiny{\makecell{Hos\_\\billing1}}&\scriptsize 77525&\scriptsize 2&\scriptsize 6&\scriptsize 217&\scriptsize 6&\scriptsize 244&\scriptsize 0.096&\scriptsize 18&\scriptsize 1&\scriptsize 22&\scriptsize 18&\scriptsize 9&\scriptsize 23&\scriptsize 1791\\\hline 
\tiny{\makecell{Hos\_\\billing2}}&\scriptsize 77525&\scriptsize 2&\scriptsize 6&\scriptsize 217&\scriptsize 8&\scriptsize 356&\scriptsize 0.05&\scriptsize 17&\scriptsize 1&\scriptsize 23&\scriptsize 19&\scriptsize 9&\scriptsize 23&\scriptsize 1790\\ \hline
\end{tabular}
\end{table}
\subsubsection{Preprocessing dimension.} 
It is believed that sometimes prefixing can remove information which will be available (in the form of data attributes) by the end of the process instance \cite{b2}. Moreover, the used encoding technique provides the predictive model with different set of features. As a result, the predictions will be influenced by the characteristics of those encoding techniques. Examples of encoding techniques characteristics may include information loss, or applicability with shorter prefixes in case of index encoding. The obligation of either using shorter traces or traces with gaps is a result of the latter characteristic. 

For bucketing traces of chosen event logs, we apply single bucketing and prefix-length bucketing, with a gap of 5 events. Moreover, we apply aggregation-based and index-based encoding techniques. Normally, both encoding techniques are coupled with static encoding to transform static attributes to be input to the predictive model along with their dynamic counterparts. Two combinations of bucketing and encoding techniques are applied to satisfy both preprocessing tasks. These combinations are \emph{single-aggregation} and \emph{prefix-index}. As index encoding leads to dimensionality explosion due to encoding each categorical level of each feature as a separate column, we decided to apply this encoding on certain event logs, i.e., \emph{Sepsis}(the three derived event logs), \emph{Traffic\_fines} and \emph{BPIC2017\_Refused}. 
\subsubsection{ML model dimension} 
In this study, we are interested in explaining the predictions of process instances outcomes. This type of prediction task is a binary classification task. In this study, we employ two predictive models, i.e., XGBoost and Logistic regression (LR). Our target is to study the influence of different transparency degrees offered by these models on explainability (\textbf{RQ3}), while predicting outcomes for inputs with varying characteristics. 

\subsubsection{XAI dimension}
In terms of XAI methods, we apply Permutation Importance, and SHAP as global XAI methods. These methods provide visual and numerical explanations, while weighting features based on their importance to the predictive model. Our choice of these XAI methods comes as a complement to querying LR for weights of its predictors and XGBoost for the most important features.

\subsection{Experiments description} \label{expdesc}
Referring back to the taxonomy from Figure \ref{ExpTax}, we carry out different experiments at each PPM task dimension. For preprocessing and ML model dimensions, experiments conducted are related to analysing the influence of using different techniques (in case of the preprocessing dimension) and the model employed (in case of the ML model dimension). 

\subsubsection{Data dimension}
The intention of analysis carried out in this dimension is to pinpoint data characteristics and relations that have the potential of affecting the patterns that were learned by a predictive model. Analysis results from this dimension are regarded when explaining model predictions in order to study the effects of observed characteristics and relationships between features. Analysis in this dimension includes:
\begin{itemize}
    \item {\emph{Correlations analysis.}} Correlation coefficient is a measurement used to describe the degree to which two variables are linearly related \cite{MLBook}. It takes values between -1 and 1 with higher values indicating a stronger relation. The sign represents the direction of the relation. Correlation coefficient is computed for normalised values of the features to be investigated. In case of original event logs, we computed correlations between categorical attributes before encoding them with Cramer's V coefficient \cite{cramercorr}. The latter measures the correlation between two nominal variables and is based on Pearson's chi-square.
    \item {\emph{Mutual Information (MI).}} Correlations between features are indicators of the degree of dependency. However, correlations are not decisive with respect to independence between features \cite{MLBook}. This means that if two variables are independent, their correlation coefficient equals to zero. However, note that a correlation coefficient of zero does not indicate independence of two variables \cite{MLBook}. We measured MI of features with respect to the label. MI is the reduction of uncertainty of a variable after observing the dependent one \cite{MLBook}. MI is capable of capturing any kind of dependencies unlike F-test, which captures only linear dependency \cite{scikitlearnCompare}. MI takes values between 0 and 1, with 0 meaning that the label is independent from the feature and 1 meaning that they are totally dependent. 
    \item {\emph{Profiling an event log.}} For each event log, before and after preprocessing, we generated a statistical profile called \emph{pandas profile} \cite{pandasprof}. Each pandas profile reports on statistical characteristics of each feature of the event log. Such characteristics include, for example, descriptive statistics of the feature, quantile statistics, missing values, most frequent values, and histograms.
\end{itemize}

Experiments in the data dimension are applied on the original event logs before the preprocessing step of the PPM workflow, i.e., before bucketing and encoding an event log. Afterwards, the same experiments are repeated on the transformed event logs. The latter is done to analyse the effect of increased dimensionality of an event log after features encoding on relationships between latent features. In addition, these experiments are conducted to study which levels in case of categorical attributes are highly correlated to other attributes. 

\subsubsection{XAI dimension}
Explainability experiments are divided into two sets, experiments depending on intrinsic explainability, and experiments depending on post-hoc XAI methods. Intrinsic explainability is achievable through mechanisms provided by the predictive model to query it for features influencing predictions generation. We compare explainability of predictive models to results generated by each applied XAI method. 

Permutation Feature Importance, and SHAP (the global form) are the model-agnostic methods we used. Moreover, we tend to analyse Logit coefficients and important features to XGBoost, as a result of having Logit and XGBoost being the predictive models used in the context of this research. To check stability of executions, we run the whole experiments taxonomy with different settings twice. Stability checks following this definition are expensive to run, due to expensive computational costs. These costs are affected by the number of datasets with different sizes (where some of the datasets experience dimensionality explosion after the preprocessing phase, and this complicates subsequent explainability steps), and the number of explainability methods applied. As a result, running the whole experiments taxonomy in the context of stability checks was not possible to be done for more than twice. Maybe depending on smaller event logs in the future enables more systematic stability check over higher number of runs. 

\section{Results and Observations} \label{6}
It is important to view explanations in the light of all contributing factors, e.g., input characteristics, the effect of preprocessing inputs, and the way how certain predictive model characteristics affect its reasoning process. In this section, we illustrate the observations we made during the experiments defined in Section \ref{5}. Due to lack of space, we focus on the most remarkable outputs illustrated by figures and tables. Further results can be generated by running the code of the experiments, which can be accessed via our Github repository.

\subsection{Data- and Preprocessing-related observations} \label{DPObserv}
In the data analysis experiments we performed before preprocessing the event logs, we made different observations. We believe that these observations are fundamental to understand model behavior. Moreover, in the data analysis experiments we carried out after preprocessing the event logs, we could observe the effects of preprocessing techniques used and formed expectations in respect to predictive model's behavior. These expectations were either confirmed or contradicted by observations we made in experiments related to other dimensions. Studying the data characteristics before and after preprocessing the logs, revealed a number of interesting observations:

\noindent\fbox{\parbox{\textwidth}{%
(1) Several features that had a constant value before encoding, turned to have multiple values after preprocessing as an effect of the applied encoding technique.}}\\
    
\noindent For example, in the \emph{BPIC2017\_Refused} event log, the feature "EventOrigin" has one value "other" before preprocessing, but turned to have 20 values after applying frequency aggregation encoding. Moreover, some features remained constant even after encoding, for example, "case:ApplicationType" in the same event log after being encoded using frequency aggregations. Some aggregation functions resulted in constant valued features, e.g., "min"-extracted features in all event logs. In the same time, index encoding technique has isolated categorical levels in separate columns and resulted sometimes in constant features. For example, in the \emph{Traffic\_fines event} log, "event\_nr" has constant columns representing each value found originally for this feature in the original event log before encoding.\\
\noindent\fbox{\parbox{\textwidth}{%
(2) Some features show a high percentage of zeros before and after encoding.}} \\

\noindent For example, in \emph{Traffic\_fines}, "points" had about 97\% of zeros before encoding. This percentage is the same for  both encoded versions of the same event log. However, zeros percentages are escalated in almost all event logs after preprocessing, as a result of applying one-hot encoding in index encoding, and counting frequencies of categorical levels in aggregation encoding. Zeros percentages reached 99\% in index-based encoded event logs, e.g., \emph{Traffic\_fines}. Moreover, in aggregation-based encoded event logs, some event logs have relevantly-low percentage of zeros,e.g., \emph{Sepsis2} with about 33\% of zeros; whereas others have high percentage of zeros between 44-99\%, e.g., \emph{BPIC\_2017}. This observation can be justified by two factors:
    \begin{itemize}
        \item {\emph{The effect of a bucketing technique.} Aggregation encoding is combined with single bucketing, which buckets all prefixes in the same group. It is more likely that when having many prefixes of the same process instance reduces the effect of feature values imbalance. Such feature values imbalance might happen based on the presence of prefixes generated from longer process instances. Moreover, index encoding is combined with prefix bucketing, which reduces the number of process instances fed into each encoding technique. As a result, combining index encoding with prefix-based bucketing has the potential of magnifying imbalances in feature values.}
        \item {\emph{The difference in 'zero' indication in both encoding techniques.} In aggregation encoding, a zero means that the feature did not have value in the encoded event. Furthermore, a zero in index encoding means whether the feature has value in the whole process instance. Note that after aggregation encoding, a process instance might be represented along many rows, whereas after index encoding, it is represented by only one row. As a result, a high number of zeros do not implicate the absence of a value for the feature after the execution of some events in aggregation encoding. A high number of zeros after index encoding denotes the absence of a value. Whatever the reason, the fact is that a certain feature will be fed into a predictive model while containing a large percentage of zeros.}  
    \end{itemize}
\noindent\fbox{\parbox{\textwidth}{%
(3.1) MI, i.e., the dependency between features and the true label in event logs, has low values in most cases.}}\\

\noindent An exception to this observation holds in the case of single aggregated version of \emph{BPIC2017} logs. Here, the remarkable observation  is that labels in these logs depend on feature "min\_event\_nr" which remains constant along the event log among the top five highly dependent features with the label. In addition, these single-aggregated logs have a high dependency between aggregated forms of "timesincecasestart" feature, which is the most dependent feature on the label in the original form of these logs. The same observation can be made in the single-aggregated versions of \emph{Sepsis2} and \emph{Sepsis3}, whereas "remainingtime" depends on the label with a coefficient of nearly 0.4, and reflecting similar dependency in the original log with lower coefficient.\\
\noindent\fbox{\parbox{\textwidth}{%
(3.2) The dominance of categorical levels of certain categorical attributes in prefix-indexed versions of some event logs. Such dominance increases with the length of the prefixes.}}\\

\begin{figure}
        \centering
        \includegraphics[width=\textwidth, height=\textheight]{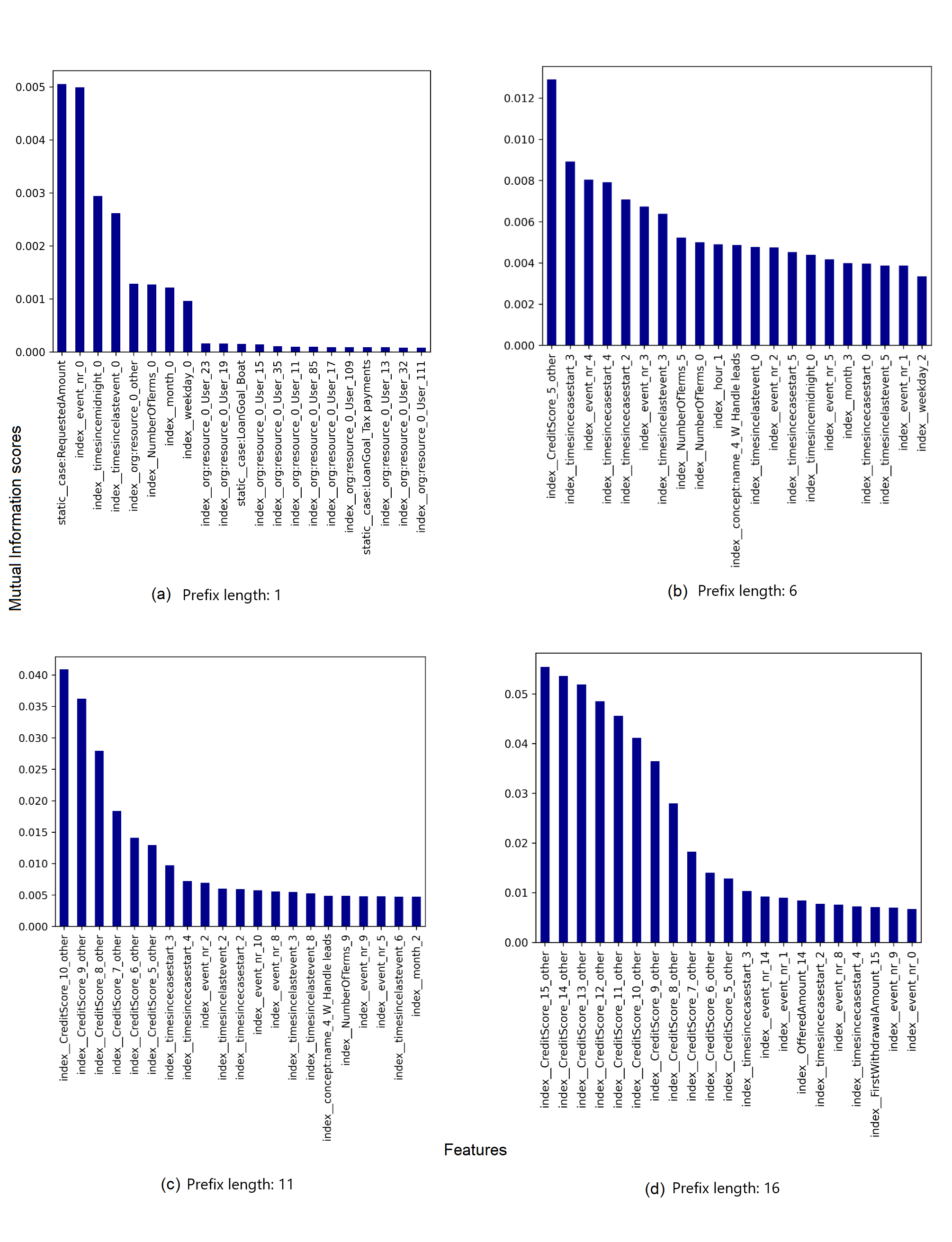}
        \caption{Mututal Information analysis of \emph{BPIC2017\_Refused} preprocessed with prefix index combination}
        \label{BPICRefusedMI}
    \end{figure}

\noindent This observation holds in the dominance of categorical levels of the "credit\_score"  feature in the prefix-indexed version of \emph{BPIC2017\_Refused} logs (cf. Figure \ref{BPICRefusedMI}). The same observation can be made for the prefix-indexed version of \emph{Sepsis3}. 

\noindent Studying correlations between features uncovers the following observations:

\noindent\fbox{\parbox{\textwidth}{%
(4.1) Two features are completely correlated across all event logs before and after applying the two preprocessing combinations.}}\\

\noindent These features are "hour" and "timesincemidnight". Note that these two features are artificial and their correlation is normal as they are both extracted from the same original DateTime column.\\
\noindent\fbox{\parbox{\textwidth}{%
(4.2) Some domain-related features inherited correlations from the original logs and propagated this correlation in a strong form in the encoded versions of the same event log.}}\\

\noindent For example, "monthlycost", "numberofterms", "offeredamount", and \\"requestedamount" are strongly correlated in the three versions of the \emph{BPIC2017} event logs. After applying single aggregation preprocessing, the same correlations remained, followed by correlations between aggregated versions of computed attributes like "timesincemidnight",  "timesincelastevent", "timesincecasestart". The same happens after applying prefix index preprocessing, with longer prefixes suffering from multicollinearity between category levels of the same attribute, e.g., "org:resource". \emph{Hospital\_billing} event logs, which are preprocessed with single aggregation combination, also inherited correlations between categorical features from the original logs. In addition, partial correlations appear between numerical features after preprocessing the original logs.

\noindent Another example of this observation is present in the complete correlations between categorical levels of certain features, e.g., "org:resource" and "dismissal", along with many other high correlations. These high correlations appear in event logs of longer prefixes in \emph{Traffic\_fines} more frequently than event logs of shorter prefixes. 

\begin{sidewaysfigure}
            \centering
            \includegraphics[width=0.9\textwidth, height=0.65\textheight]{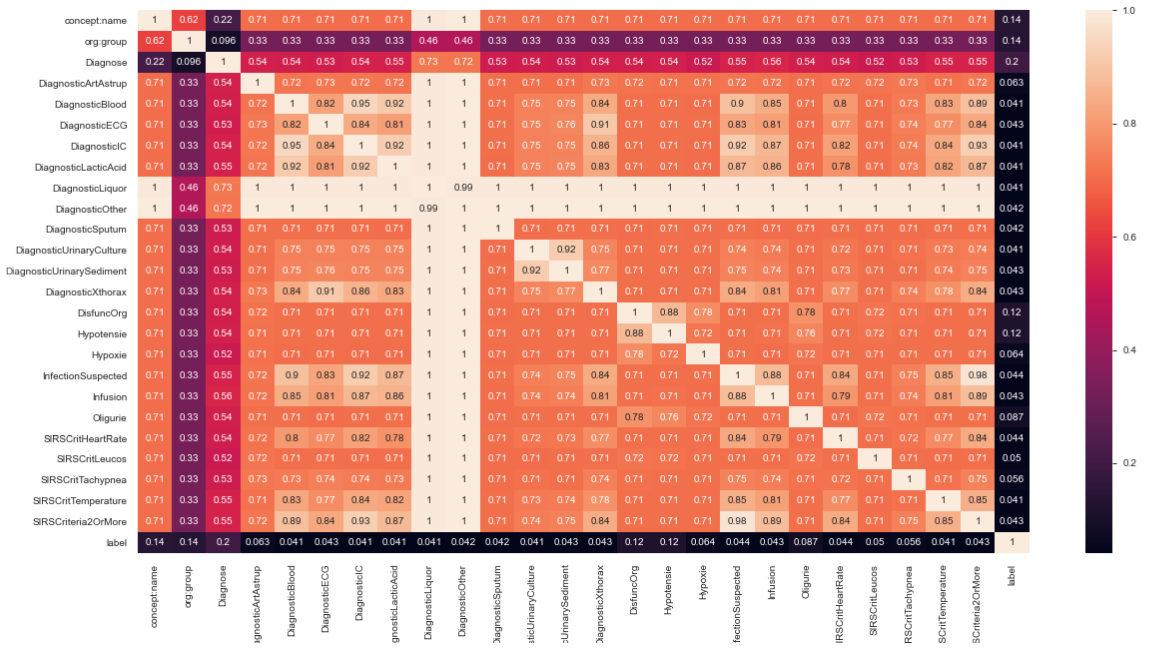}
            \caption{Collinearity between categorical attributes in \emph{Sepsis2}} \label{corrcatsepsis2beforeenc}
\end{sidewaysfigure}

\noindent The most clear example is manifested in the high collinearity between categorical attributes in original \emph{Sepsis} logs (as shown in Figure \ref{corrcatsepsis2beforeenc}) is inherited as complete collinearity in the event logs we obtain after applying either preprocessing combination. However, the difference becomes obvious for numerical features in both preprocessing combinations. Partially, high correlations exist between aggregated forms of numerical features in single-aggregated event logs. Finally, only few correlations exist between numerical features in shorter prefixes and increase in number when prefixes get longer.

\noindent\fbox{\parbox{\textwidth}{%
(4.3) High correlations emerge between features that are not correlated in the original log, while highly correlated after applying both preprocessing combinations.}}\\

\noindent As an example consider the correlations in \emph{Traffic\_fines} event logs after preprocessing, where some of the correlated features contain constant values.
\subsection{XAI-related observations} \label{exmlobs}
\subsubsection{Observations of model-specific explanations} \label{obsmodelspec}
In the experiments described in Section \ref{expdesc}, we apply LR and XGBoost. Both provide some insights into the importance of features used in the prediction task, through coefficients and features importance in LR and XGBoost, respectively. As we will discuss later in this section, these methods are not always providing reliable results. However, the results can indicate which features have influenced the predictive model decisions and their order, regardless of the weight of these features. 

Interpreting features coefficients or weights in the context of LR for example, is not a straight forward process the way it is in linear regression. LR employs a log function and, hence, outputs the probability of belonging to a certain class \cite{MLBook}. Therefore, weights of LR do not influence the prediction linearly. Instead, a change to a feature by one unit, results in a change to the log odds ratio by the value of the corresponding weight of this feature \cite{imlbook}. Log odds is the logistic function of the probability of an event divided by the probability of no event \cite{imlbook}. 
When running LR over the used event logs twice, the following observations were made:

\noindent\fbox{\parbox{\textwidth}{%
(1) Event logs preprocessed with single aggregation combination show a 100\% matching in features coefficient analysis with respect to feature set and weights of importance. However, the former logs do not show any similarity between the most important features according to the model and the features with high dependency on the label according to MI analysis.}}\\

\noindent The \emph{Traffic\_fines} event log shows an exception of the latter observation, where it shows similarity between the first two features with the highest coefficients and the features showing high dependency with the label according to MI. According to correlations analysis, a high collinearity between these two features is observed, as they constitute aggregations of "timesincelastevent" feature. \\
\noindent\fbox{\parbox{\textwidth}{%
(2) Event logs preprocessed with prefix index combination show similarities in features coefficient analysis in terms of feature set as the length of the prefix increases. However, the weights remain different across several execution runs.}}\\

\noindent Moreover, the similarity between most important features according to the model and the set of features with high dependency with the label increases with prefix length increases, as well. An exception to both observations on prefix indexed- event logs is the \emph{Traffic\_fines} event log. The non-increasing similarity between feature sets across execution runs on this event log can be justified with the note that this log only provides shorter prefixes (cf. Table \ref{statistics}). Therefore, there is no chance to observe a change on longer traces unlike in other event logs.

Using the built-in capability of XGBoost to query the most important features used by the model can provide feature sets ranked or weighted based either on \emph{gain}, \emph{weight}, and \emph{cover} according to XGBoost API documentation \cite{XGBoostAPI}. Two complementary importance criteria are available through the scikit-learn implementation of XGBoost. These criteria are \emph{total\_gain}, and \emph{total\_cover}.

\emph{Gain} is considered to be the most relevant attribute to measure the relative importance of each feature, especially as MI is the measurement representing the expected gain from a data analysis perspective. Comparing results from both a data analysis perspective and a model reasoning perspective, and observing how both sides (dis)agree result in interesting insights. These insights highlight how a predictive model learns from the underlying data and reflects their anticipated characteristics. However, in our experiments, we have analysed features importance results based on all five criteria. Note that in our observations there are two usages of the term \emph{weight}. There is \emph{weight} indicating  one of the five feature importance criteria analysed, and the \emph{weight} of a feature in the important feature set according to one of the five importance criteria. The following observations are made as a result of analysing features importance according to XGBoost-based models:

\noindent\fbox{\parbox{\textwidth}{%
(1) XGBoost models show total inconsistency across the two execution runs for all event logs preprocessed with single aggregation combination.}}\\

\noindent The three \emph{Sepsis} event logs constitute an exception. They have the same feature set with different weights for four of the criteria, whereas the feature set increasing the gain of the model is totally different across the two execution runs. However, in event logs preprocessed with prefix index combination, models trained on prefix length-based event logs use the same feature sets with the same weights across the two execution runs for the five criteria. An exception of this consistency is observed in prefix length-based event logs extracted from the \emph{BPIC2017\_Refused} event log.\\
\noindent\fbox{\parbox{\textwidth}{%
(2.1) Multicollinearity, i.e., high correlations reaching complete correlations in some cases, is an issue in event logs preprocessed with single aggregation combination.}}\\

\noindent An exception of this observation is made on the three event logs extracted from the \emph{BPIC2017} log. This multicollinearity can be observed across the feature set of each criterion in the same execution run. Moreover, It can be observed in feature sets of the same criterion across the two execution runs.

\noindent\fbox{\parbox{\textwidth}{%
(2.2) This multicollinearity is not observed in event logs preprocessed with prefix index combination}}\\

\noindent An exception to the latter observation is present in the three \emph{Sepsis} event logs and their prefix length-based extracted event logs. This can be justified by the fact that \emph{Sepsis} logs inherit very high correlations from the original \emph{Sepsis} logs (cf. Section \ref{DPObserv}). \\
\noindent\fbox{\parbox{\textwidth}{%
(3) Feature sets which are important according to their gain differ from the ones with high dependency on the label as indicated by the MI analysis.}}\\

\noindent This observation applies to all event logs (independent from whether being preprocessed with single aggregation or prefix index combination) across the two execution runs, except for the prefix length-based event logs extracted from the \emph{BPIC2017\_Refused} event log. For the later event logs, gain-based feature sets tend to match the MI-based feature sets, with increasing length of the prefixes in event logs.\\
\noindent\fbox{\parbox{\textwidth}{%
(4) In the event logs preprocessed with prefix index combination, with increasing length of the prefix, feature sets that are considered being important according to their weight and their cover importance criteria tend to increase in their similarity across event logs.}}\\

\noindent However, for similar features across the same criterion, the weight of the feature importance decreases in longer prefixes. \emph{Traffic\_fines} prefix length-based event logs provide an exception of this observation. This exception can be justified by the fact that these logs do not provide longer prefixes when applying a gap of 5 events (see the experiments settings in Section \ref{ExpSetup}).

\begin{figure}
    \centering
    \includegraphics[width=1\textwidth]{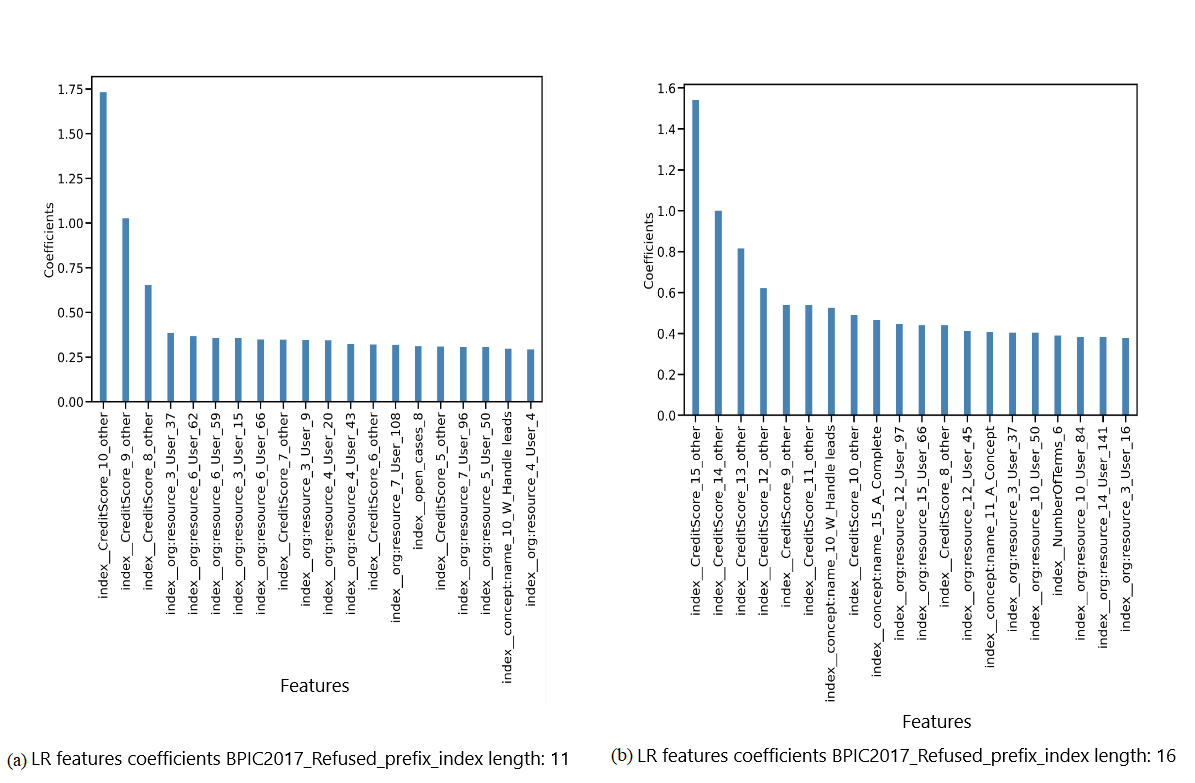}
    \caption{LR coefficients for different prefix lengths of \emph{BPIC2017\_Refused}}
    \label{LRcoefBPICRefusedprfxidx}
\end{figure}

\begin{figure}
    \centering
    \includegraphics[width=\textwidth, height=0.4\textheight]{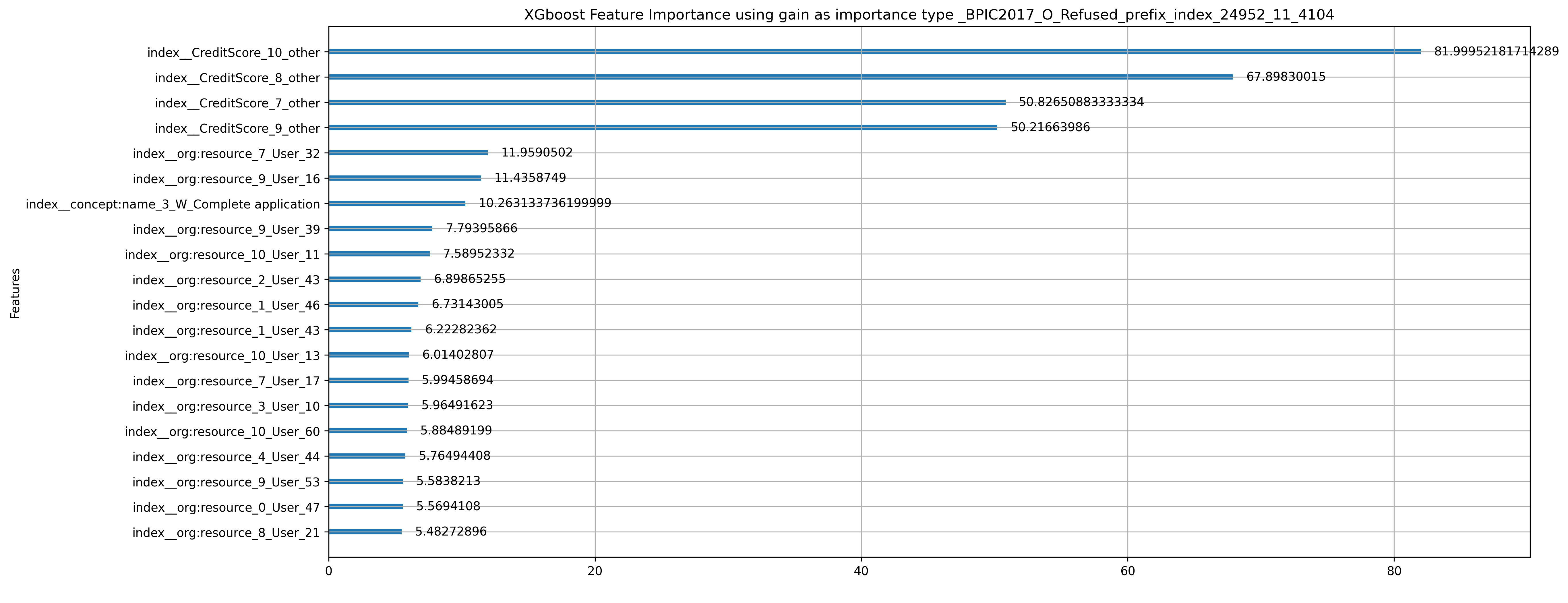}
    \caption{XGBoost features importance (gain) for prefix length (11) of \emph{BPIC2017\_Refused}}
    \label{XGboostGainBPICRefusedprfxidx_11}
\end{figure}

\begin{figure}
    \centering
    \includegraphics[width=\textwidth,height=0.4\textheight]{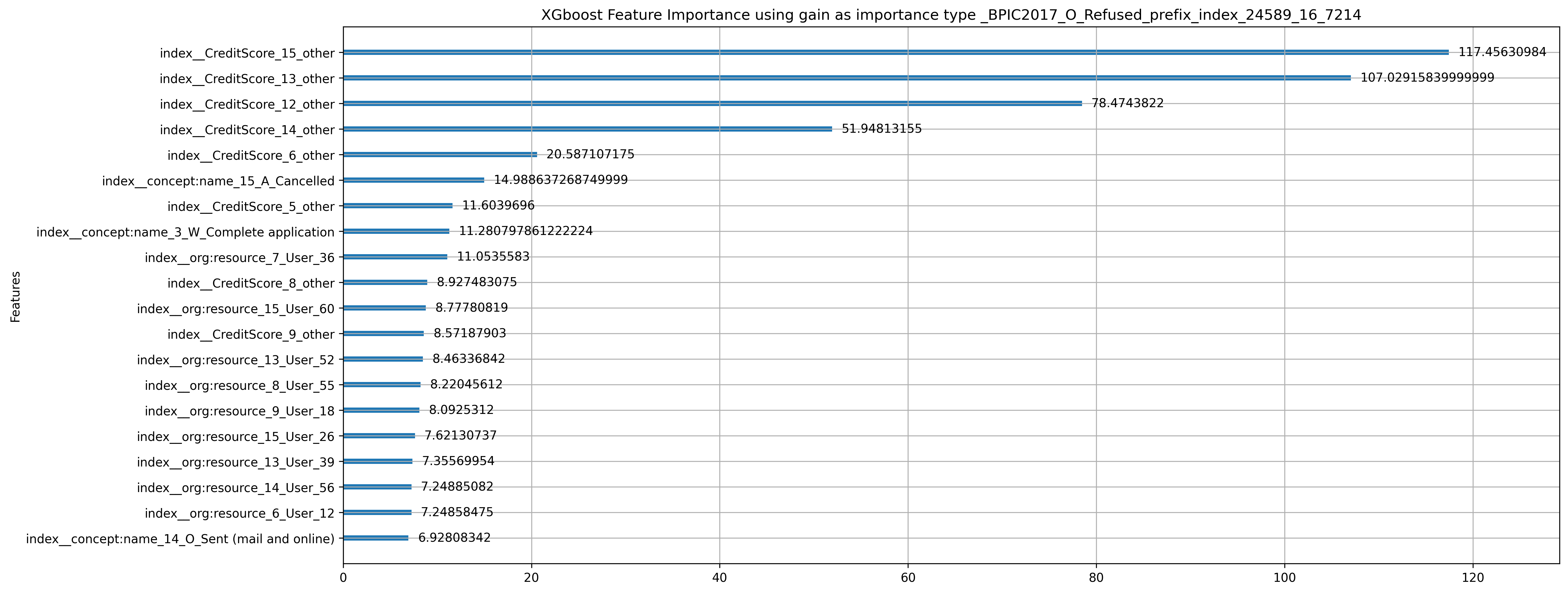}
    \caption{XGBoost features importance (gain) for prefix length (16) \emph{BPIC2017\_Refused}}
    \label{XGboostGainBPICRefusedprfxidx_16}
\end{figure}

For example, Figures [\ref{LRcoefBPICRefusedprfxidx}-\ref{XGboostGainBPICRefusedprfxidx_16}] compare LR coefficients with XGBoost features importance (based on gain criterion). The comparison indicates increasing similarity in the important features subset with increasing prefix length. 
\subsubsection{Observations of global XAI methods results} \label{obsglobexml}
In this subsection, an analysis of Permutation Feature Importance, and SHAP results is presented. For each XAI method, we search for indications of multicollinearity and (dis)conformance to important feature sets indicated by the employed predictive model; whether being a LR or XGBoost; and other interesting observations.
\paragraph{\textbf{Permutation Feature Importance (PFI).}} The basic idea of PFI is to measure the average between the error in prediction after and before permuting the values of a feature \cite{imlbook}. Feature values permutation or shuffling aims to estimate the increase in prediction error as an indicator of the feature importance. PFI is executed twice to query LR and XGBoost models trained over the event logs preprocessed with single aggregation and prefix index combination. Each execution run included 10 shuffling iterations. The mean importance of each feature is computed. In detail, PFI execution led to the following observations:\\
\noindent\fbox{\parbox{\textwidth}{%
(1) In single-aggregated event logs, the results of the two runs are consistent with respect to feature sets and the weights of these features. However, the resulting feature sets are affected by multicollinearity between the features. In prefix-indexed event logs, the two runs are consistent in all event logs.}}\\
\begin{figure}
        \centering
        \includegraphics[width=\textwidth]{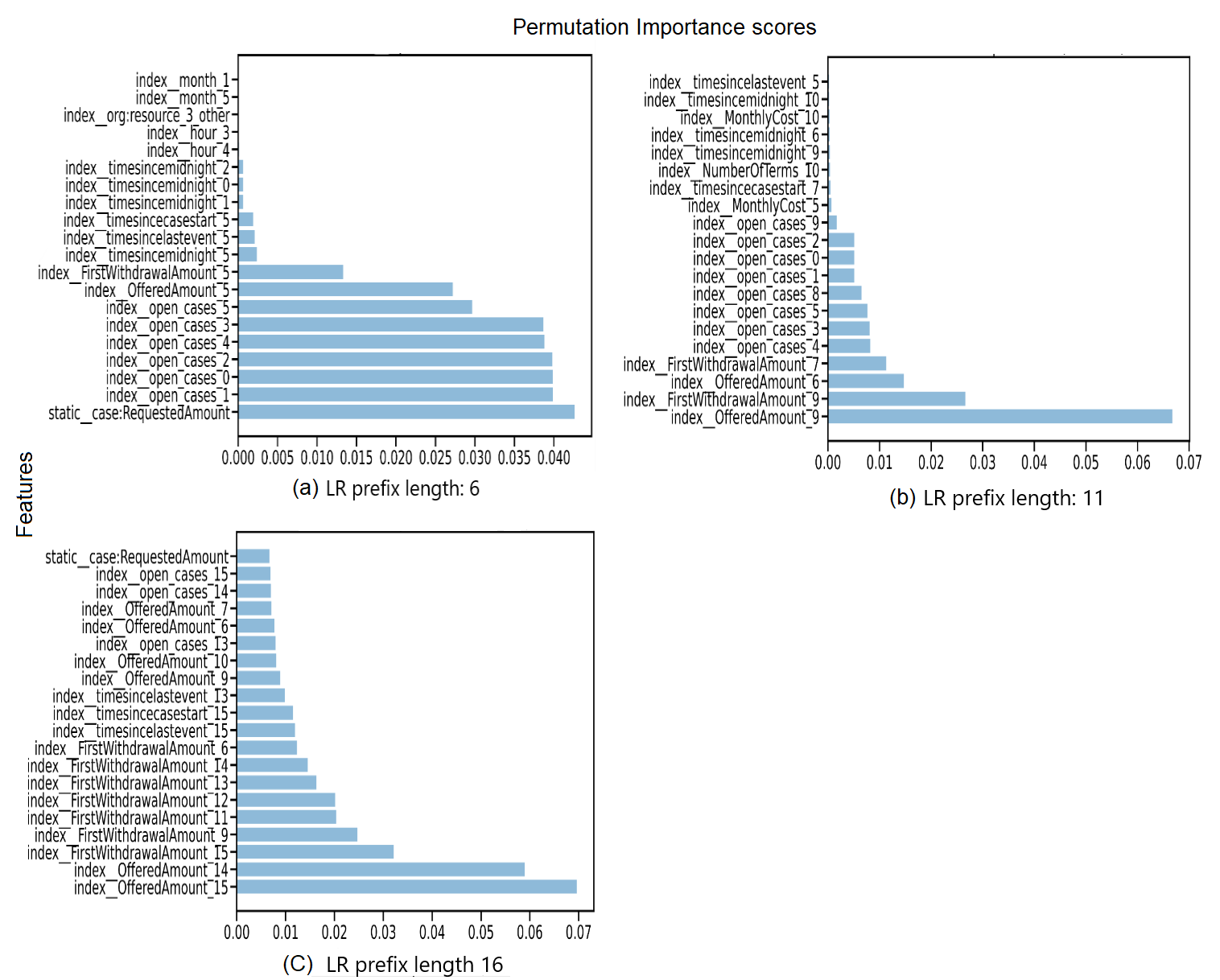}
        \caption{PFI according to LR trained over \emph{BPIC2017\_Refused} (preprocessed with prefix index combination)}
        \label{BPICrefpfiLR}
\end{figure}
\begin{figure}
        \centering
        \includegraphics[width=\textwidth, height=\textheight]{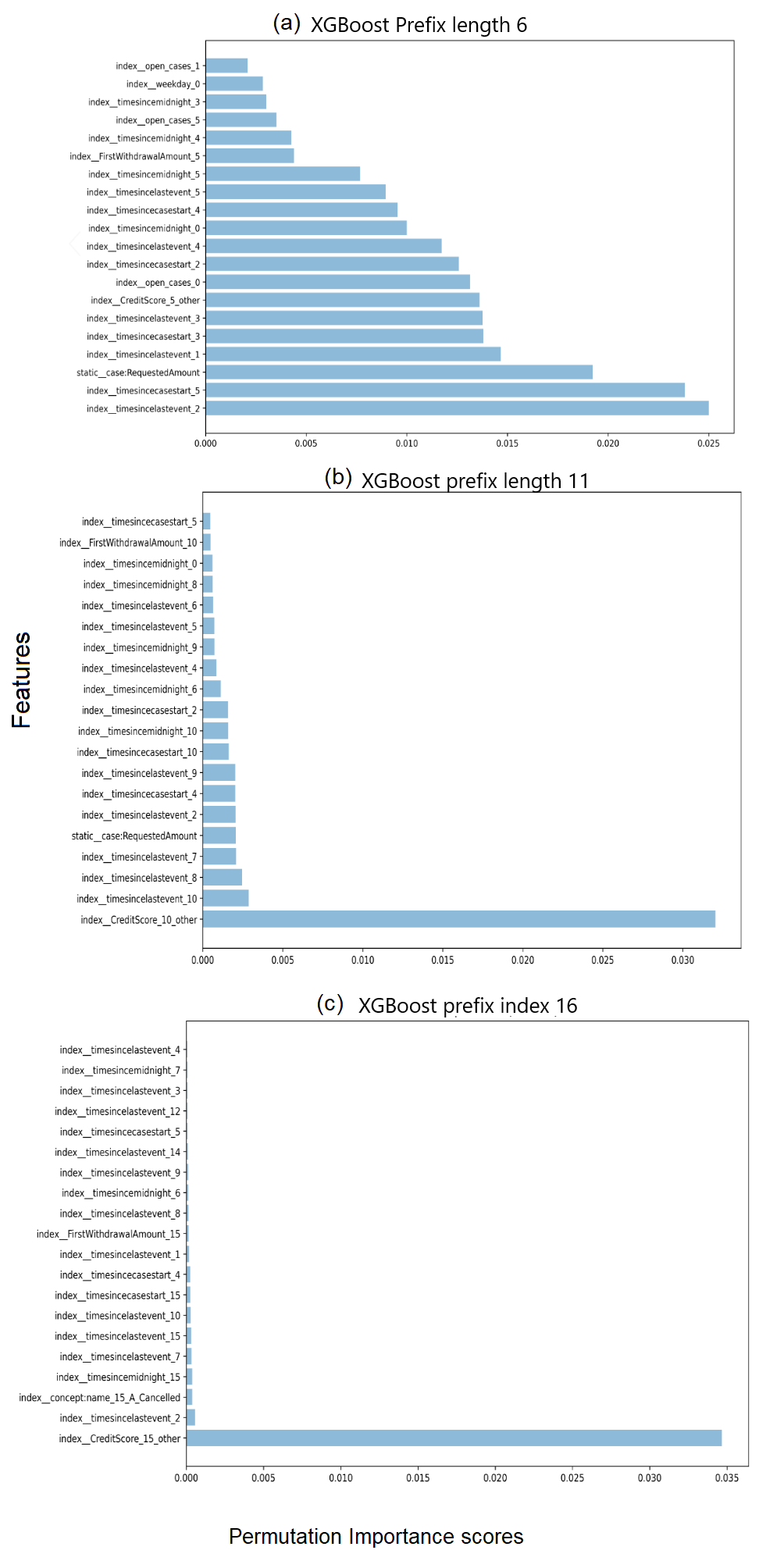}
        \caption{PFI according to XGBoost trained over \emph{BPIC2017\_Refused} (preprocessed with prefix index combination)}
        \label{BPICrefpfiXGBoost}
\end{figure}
\noindent An exception in prefix-indexed event logs is present in ones derived from \\\emph{BPIC2017\_Refused}. In the latter event logs, the dissimilarity between the feature sets across the two runs increases with increasing length of the prefixes. This observation can be attributed to the effect of the increased dimensionality in the event logs with longer prefixes. For prefix-indexed event log, weights of important features change with increasing prefix length.\\
\noindent\fbox{\parbox{\textwidth}{%
(2) (Dis)similarity between PFI's first run on one hand and LR coefficients or XGBoost feature sets on another, in independent of the used preprocessing combination used. In nearly all cases, dissimilarity can be observed.}}\\

\noindent After comparing the results of the first PFI run with LR coefficients and XGBoost feature sets (based on the total gain criterion), complete dissimilarity could be observed between the PFI feature sets and LR coefficients, regardless the preprocessing technique used. Moreover, when compared to XGBoost feature sets, event logs showing no similarity in a single-aggregated form, and start to show more similarity in some features in prefix-indexed form with increasing length of the prefixes. Examples of this observation include the \emph{BPIC2017\_Refused}, \emph{Sepsis1}, \emph{Sepsis2} event logs. Figures \ref{BPICrefpfiLR} and \ref{BPICrefpfiXGBoost} show PFI scores according to LR and XGBoost trained over \emph{BPIC2017\_Refused} of different prefix lengths. \\
\noindent\fbox{\parbox{\textwidth}{%
(3) Multicollinearity between PFI feature sets of all single-aggregated event logs is at high levels, while it increases as the length of prefixes increase in prefix-indexed event logs.}}\\

\noindent Multicollinerity implies having features with very high or complete correlations among each other.
\paragraph{\textbf{SHapley Additive exPlanations (SHAP).}}
SHAP is an explanation method belonging to the class of feature additive attribution methods \cite{shapL}. These methods use a linear explanation model to compute the contribution of each feature to a change in the prediction outcome with respect to a baseline prediction. Afterwards, a summation of the contributions of all features approximates the prediction of the original model. 
To maintain comparability of the global XAI methods used in our experiments, we constructed a SHAP explainer model on training event logs independently of another SHAP explainer model constructed on relevant testing event logs. The concluded observations made in this section are drawn based on the training SHAP explainer model.\\
\noindent\fbox{\parbox{\textwidth}{%
(1) While comparing the two execution runs, results did not depend on the preprocessing technique used, but differed depending on the predictive model being explained.}}\\

\noindent While explaining predictions of the LR model, performing two executions of the SHAP method did neither result in different feature sets nor different ranks based on SHAP values, regardless the preprocessing combination used. Meanwhile, explaining predictions of XGBoost model reveals having the first most contributing feature as being the same across both execution runs, while the rest of the feature set is the same, but differs in features ranks. An exception is present in the feature set of the three \emph{Sepsis} event logs, where feature ranks are the same across both runs. These observations are aligned with the observations made regarding model-specific explanations and features importance according to XGBoost-based models across two execution runs.

\noindent To study how SHAP results relate to feature importance as revealed by the predictive models, it becomes necessary to compare the features ranked according to their SHAP values with the important feature sets as indicated by predictive models used in these experiments.\\
\noindent\fbox{\parbox{\textwidth}{%
(2.1) Important features as being ranked by their coefficients in LR are unaligned with important features being ranked by their SHAP values.}}\\ 

\noindent This observation is valid across all event logs regardless of the used preprocessing technique.\\
\noindent\fbox{\parbox{\textwidth}{%
(2.2) While using the total gain as a criterion to rank important features according to XGBoost models, the former observation is not valid.}}\\

\noindent In prefix-indexed event logs, there is some similarity between the two compared feature sets. Meanwhile, in single-aggregated event logs, only in \\ \emph{BPIC2017\_Accepted}, \emph{Hospital\_billing\_2}, \emph{Sepsis1}, \emph{Sepsis3} there exists some similarity -but not a complete match- between the two compared feature sets.\\
\noindent\fbox{\parbox{\textwidth}{%
(3) Despite adjusting SHAP explainer parameters to be true to the model rather than the underlying relations between features, SHAP results reveal high multicollinearity between the most important features}}\\

\noindent This observation holds especially in predictions of event logs preprocessed with single aggregation combination. It is valid on the feature sets important to LR models, while being valid to XGBoost models in case of \emph{Hospital\_billing\_2} and \emph{Sepsis1} event logs. Multicollinearity in the underlying data affects LR as its presence counterfacts underlying assumption of LR. Meanwhile, multicollinearity is not supposed to affect XGBoost. This claim is justified by the fact that in boosting, whenever collinearity exists between a subset of features, the model chooses one feature to be the data splitting criterion. The splitting feature is then assigned all the importance score, compared to the left out correlated features that won't be considered important in this case \cite{UnderXGBoost}. However, in case of \emph{Hospital\_billing\_2} and \emph{Sepsis1} event logs, the very high class imbalance (cf. Table \ref{statistics}) indicates that the model overfits  patterns in training data.

\noindent Moreover, in event logs preprocessed with prefix index combination, high correlations can be observed between features in the highly ranked feature sets according to LR with increasing prefix length. Multicollinearity in single-aggregated event logs is mainly inherited from high correlations present in original event logs. This multicollinearity increases with single aggregation preprocessing of an event log.   

\noindent On some event logs, LR models depend only on one feature according to SHAP values. The aforementioned logs are event logs with prefix length of one extracted from event logs \emph{BPIC2017\_Refused}, \emph{Sepsis1} and \emph{Sepsis2}. SHAP nullifies the contribution of features when they are not important to a predictive model. Therefore, for these given event logs, only the important feature has a contribution value, while other features receive zeros. 
\begin{figure}
\centering
\begin{subfigure}{.5\textwidth}
  \centering
  \includegraphics[width=1\linewidth,height=0.25\textheight]{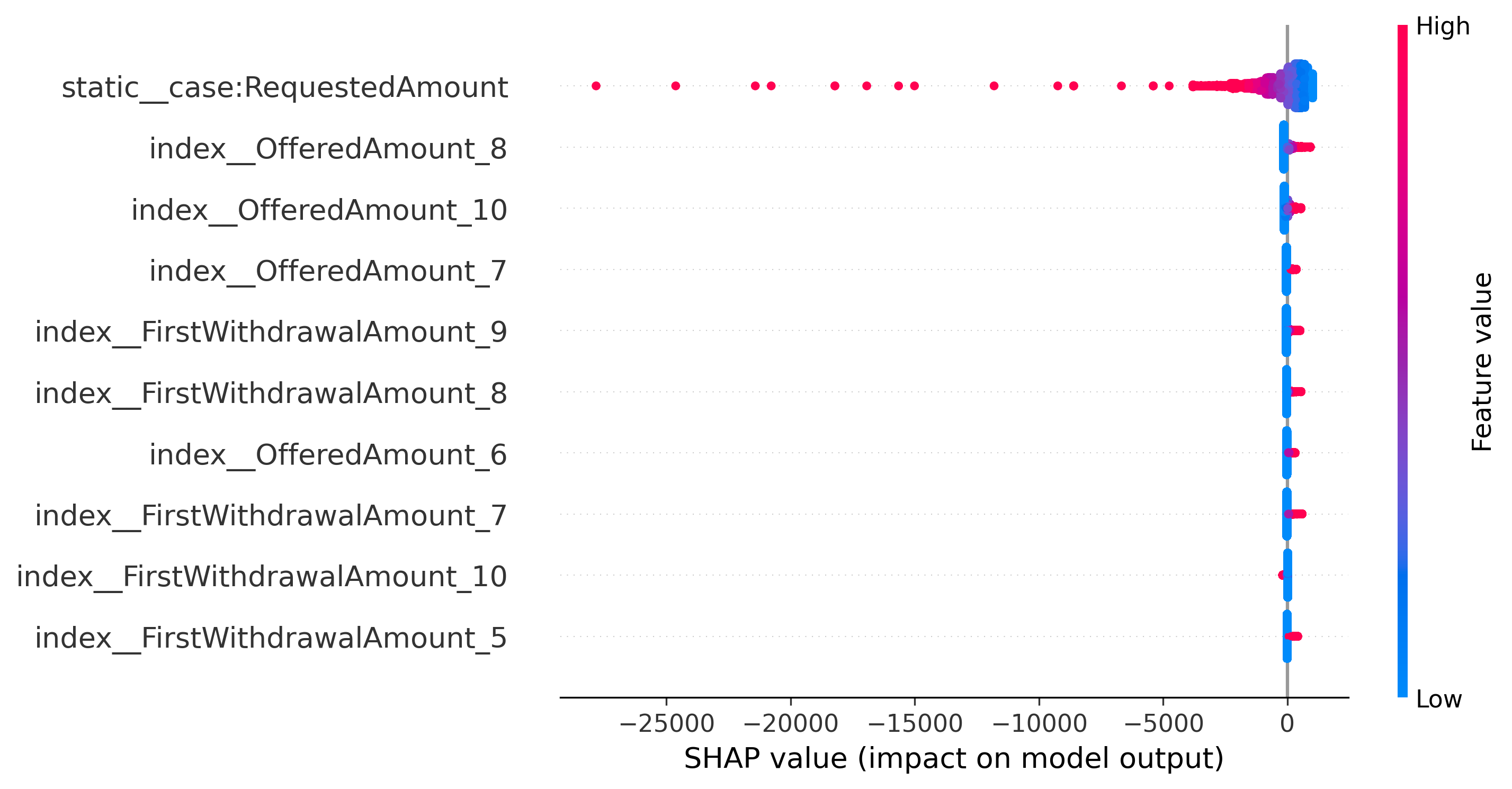}
  \caption {\tiny{SHAP values of BPIC2017\_Refused with\\prefix length of 11}}
  \label{shapBPICRef}
\end{subfigure}
\begin{subfigure}{.5\textwidth}
  \centering
  \includegraphics[width=1\linewidth,height=0.25\textheight]{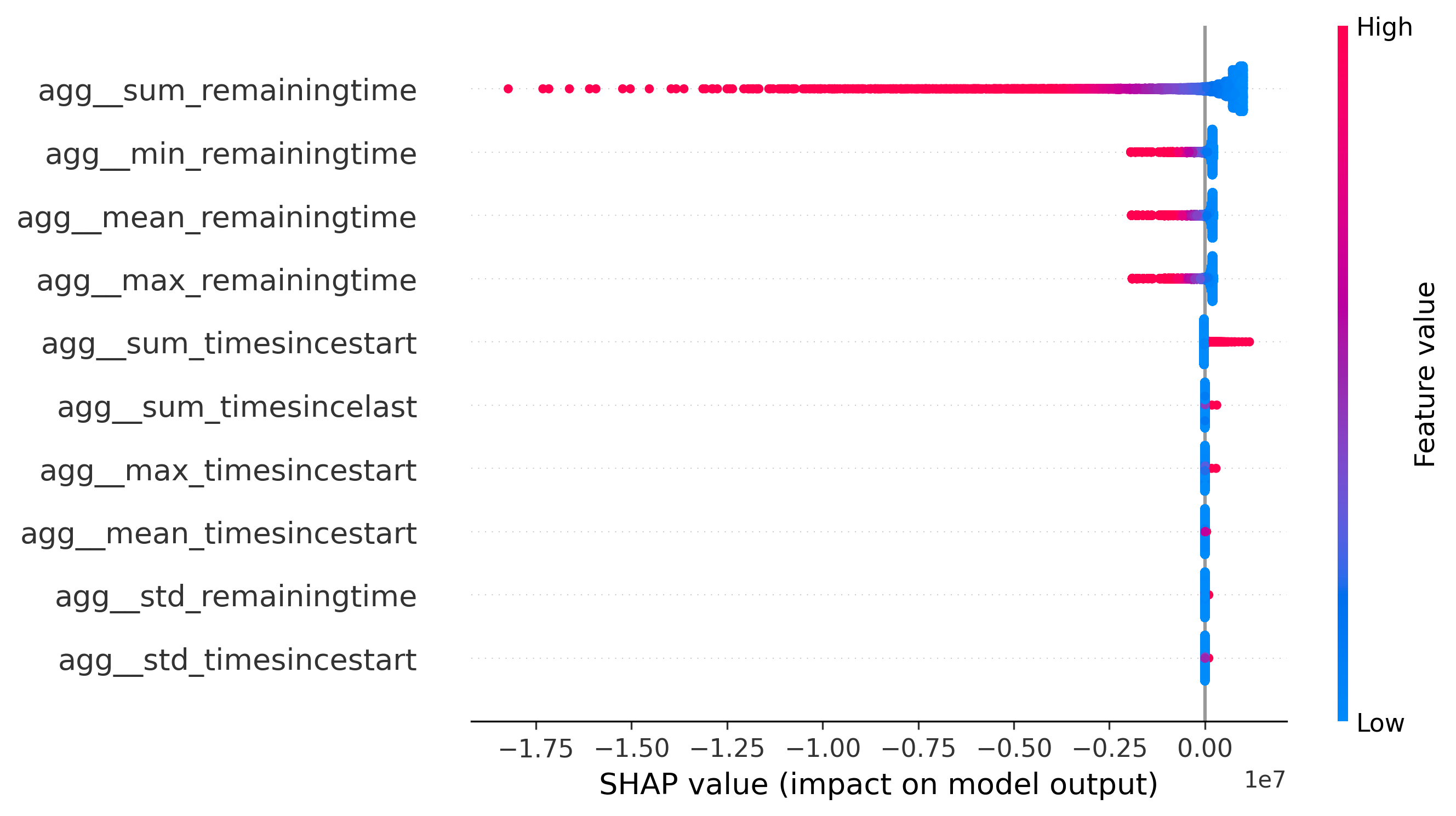}
  \caption {\tiny{SHAP values of Sepsis3 preprocessed\\using single aggregation}}
  \label{shapsepsis}
\end{subfigure}
\caption{Summary plots of SHAP values of features used by LR}
\label{shapcollinear}
\end{figure}

\noindent In Figure \ref{shapcollinear}, features whose all values are vertically allocated (around 0 in both sub-figures), are ignored by the predictive model. While features whose values are all spread along the horizontal axis, influence model predictions. This influence might be present either through lowering or increasing the values of model predictions. The color of points corresponds to feature values in relevant process instances, whether being a high or low value. 

\section{Discussion}  \label{Section 7}
Explaining ML-based predictions is a necessity for gaining user acceptance and trust in a predictive models' predictions. It is necessary to regard explainability as a continuous process which should be integrated throughout the whole ML pipeline. A first step towards such integration would be a study of the effect of different pipeline decisions on resulting explanations. Our main concern in this research is to study the ability of an explanation to reflect how a predictive model is affected by different settings in the ML pipeline. Another concern is to study how a predictive model's inherent characteristics are confirmed and highlighted through an explanation. Experiments results described and analysed in Section \ref{6} confirm the following conclusions:
\begin{itemize}
    \item Encoding techniques used in the preprocessing phase of PPM have a major impact on both the ability of the predictive model to be decisive about important features, and the selection of certain XAI methods to explain model's predictions. Both studied encoding techniques load the event log with a large number of derived features. However, the situation is worse in index-based encoding, as the number of resulting features is increasing proportional to the number of dynamic attributes, especially the number of categorical levels of a dynamic categorical attribute. In aggregation-based encoding the number of resulting features in increasing proportional to the number of dynamic attributes. 
    
    Increased collinearity in the underlying data is another problem resulting from encoding techniques with varying degrees. The effect of collinearity is observed in index-based preprocessed event logs, while not completely absent in aggregation-based event logs. This collinearity is reflected through explanations of predictions on process instances from prefix-indexed event logs as the length of a prefix increases.  

    \item Chosen bucketing technique has an effect on global XAI methods, especially when accuracy is dependent on the sufficiency of process instances to be analysed. PFI is affected by the number of process instances as the average of errors in prediction is calculated over the number of event log process instances.  
    
    \item Experiments show sensitivity of LR to collinearity in several positions. This conclusion is made for example, when comparing features ranked highly based on SHAP and PFI to high important features based on LR coefficients. Meanwhile, there is a degree of similarity when comparing the former important feature sets to XGBoost important features set, especially for explanations of predictions over prefix-index preprocessed event logs. However, while querying the XGBoost model for the set of important features twice, the resulting sets do not match, in an indication of inconsistency as a result of collinearity between features. Both predictive models are affected by collinearity. However, in LR, the effect is magnified and more prevailing in all comparisons where LR coefficients take part. In most cases, it is observed that dimensionality and collinearity draw both LR and XGBoost from relying on features which have dependency relation with the label.
\end{itemize}
This discussion emphasizes the importance of regarding explaining an event log as an accumulated effort starting from features analysis and selection stage till training a predictive model and an explanation model. Despite having XAI methods true to the model, explanations have the power of highlighting how underlying data characteristics affected and are reflected in the model reasoning process.

\section{Related Work} \label{Section 8}
Recently, awareness levels are increased about the necessity of supporting PPM results with explanations among PPM practitioners. A few number of approaches emerged to either equip the proposed PPM approach with a post-hoc explanation technique \cite{XNAP,featItems,SHapItalian}. \cite{XNAP} proposes an approach which integrates Layer-wise Relevance Propagation(LRP) to explain next activity predicted using an LSTM predictive model. This approach tends to propagate relevance scores backwards through the model to indicate which previous activities were crucial to obtaining the resulting prediction.

Another approach explaining an LSTM decisions is present in \cite{SHapItalian}. According to this approach, the total number of process instances where a certain feature is contributing to a prediction is identified at each timestamp for the whole dataset. This identification is directed by SHAP values. \cite{SHapItalian} uses the same approach in providing local explanations for running process instances. 

Explanations can also be used to leverage a predictive model performance as proposed by \cite{featItems}. Using LIME as a post-hoc explanation technique to explain predictions generated using Random Forest, \cite{featItems} identified feature sets which contributed the producing wrong predictions. After identifying these feature sets, their values are randomised, provided that they don't contribute to generating right predictions for other process instances. The resulting randomised dataset is then used to retrain the model again till its perceived accuracy is improved.

\section{Conclusions} \label{Section 10}
In this research, a framework is implemented for studying the effects of several choices made in the context of a PPM task on a ML predictive model reasoning. Using this framework, it is possible to study the ability of explanations to reflect characteristics of underlying data. Our study revealed inconsistencies between data characteristics, how a ML model uses these data, and how such usage is reflected in resulting explanations predictions. We study how explanations reflect a predictive model's sensitivity towards underlying data characteristics. 

It has also emphasized based on our experiments the importance of feature selection after preprocessing an event log, not only to maximise accuracy of the predictive model, but also to be able to produce useful explanations of predictions. Our study has highlighted situations where data problems may not affect the accuracy of predictions, but do affect usefulness and consistency of explanations. Therefore, explainability should be seamlessly integrated into PPM workflow stages as an inherent task not as a follow up effort.

\end{document}